\documentclass[letterpaper]{article} % DO NOT CHANGE THIS
\usepackage{aaai25}   % DO NOT CHANGE THIS
\usepackage{times}  % DO NOT CHANGE THIS
\usepackage{helvet}  % DO NOT CHANGE THIS
\usepackage{courier}  % DO NOT CHANGE THIS
\usepackage[hyphens]{url}  % DO NOT CHANGE THIS
\usepackage{graphicx} % DO NOT CHANGE THIS
\usepackage{natbib}  % DO NOT CHANGE THIS AND DO NOT ADD ANY OPTIONS TO IT
\usepackage{caption} % DO NOT CHANGE THIS AND DO NOT ADD ANY OPTIONS TO IT
\usepackage{algorithm}
\usepackage{algorithmic}
\usepackage{amssymb}
\usepackage{newfloat}
\usepackage{listings}
\DeclareCaptionStyle{ruled}{labelfont=normalfont,labelsep=colon,strut=off} % DO NOT CHANGE THIS
\floatstyle{ruled}
\newfloat{listing}{tb}{lst}{}
\floatname{listing}{Listing}
\title{Towards Intrinsic Self-Correction Enhancement in Monte Carlo Tree Search Boosted Reasoning via Iterative Preference Learning}

\author {
    Huchen Jiang$^*$\textsuperscript{\rm 1},
    Yangyang Ma$^*$\textsuperscript{\rm 1},
    Chaofan Ding\textsuperscript{\rm 1},
    Kexin Luan\textsuperscript{\rm 1},
    Xinhan Di\textsuperscript{\rm 1}
}
\affiliations {
    \textsuperscript{\rm 1}AI Lab, Giant Network\\
    jianghuchen@ztgame.com, mayangyang@ztgame.com, dingchaofan@ztgame.com, luankexin@ztgame.com, dixinhan@ztgame.com
}

\usepackage{bibentry}
\begin{document}

\maketitle

\begin{abstract}
With current state-of-the-art approaches aimed at enhancing the reasoning capabilities of Large Language Models(LLMs) through iterative preference learning inspired by AlphaZero, we propose to further enhance the step-wise reasoning capabilities through intrinsic self-correction to some extent. Our work leverages step-wise preference learning to enhance self-verification via reinforcement learning. We initially conduct our work through a two-stage training procedure. At the first stage, the self-correction reasoning ability of an LLM is enhanced through its own predictions, relying entirely on self-generated data within the intrinsic self-correction to some extent. At the second stage, the baseline step-wise preference learning is leveraged via the application of the enhanced self-correct policy achieved at the first stage. In the evaluation of arithmetic reasoning tasks, our approach outperforms OpenMath2-Llama3.1-8B, dart-math-mistral-7b-uniform on MATH with increases in accuracy to 71.34\%(+4.18\%) and 48.06\%(+4.94\%) and LLama-3.1-8B-Instruct, Mistral-7B-Instruct-v0.1 on GSM8K with increases in accuracy to 86.76\%(+2.00\%) and 38.06\%(+2.28\%).

\end{abstract}

\section{Introduction}

The integration of MCTS \cite{coulom2006efficient,kocsis2006bandit}, neural networks and RL techniques \cite{silver2017mastering} has been successfully developed since AlphaZero \cite{silver2017mastering} contributing to its superhuman performance across various domains. MCTS has subsequently been integrated as a policy improvement operator, transforming the current policy into an enhanced one \cite{grill2020monte}. Moreover, integrating MCTS into the iterative process of policy development could lead to significant advancements in LLMs, especially in areas such as reasoning and decision-making that align with human-like preferences \cite{zhu2022solving, hao2023reasoning}. The instance-level approach utilizes sparse supervision, which may overlook important information and fail to fully exploit the potential of MCTS in enhancing LLMs \cite{wu2023fine}. Another challenge is MCTS's dependence on a critic or a learned reward function, which is essential for providing meaningful feedback on the various rollouts generated by MCTS \cite{liu2023making}. To address this granularity issue, research in LLMs suggests that process-level or stepwise evaluations are superior to instance-level evaluations \cite{lightman2023let,li2023making,xie2023decomposition,yao2024tree,rafailov2024direct}.

\begin{figure}
    \centering
    \includegraphics[width=1.0\linewidth]{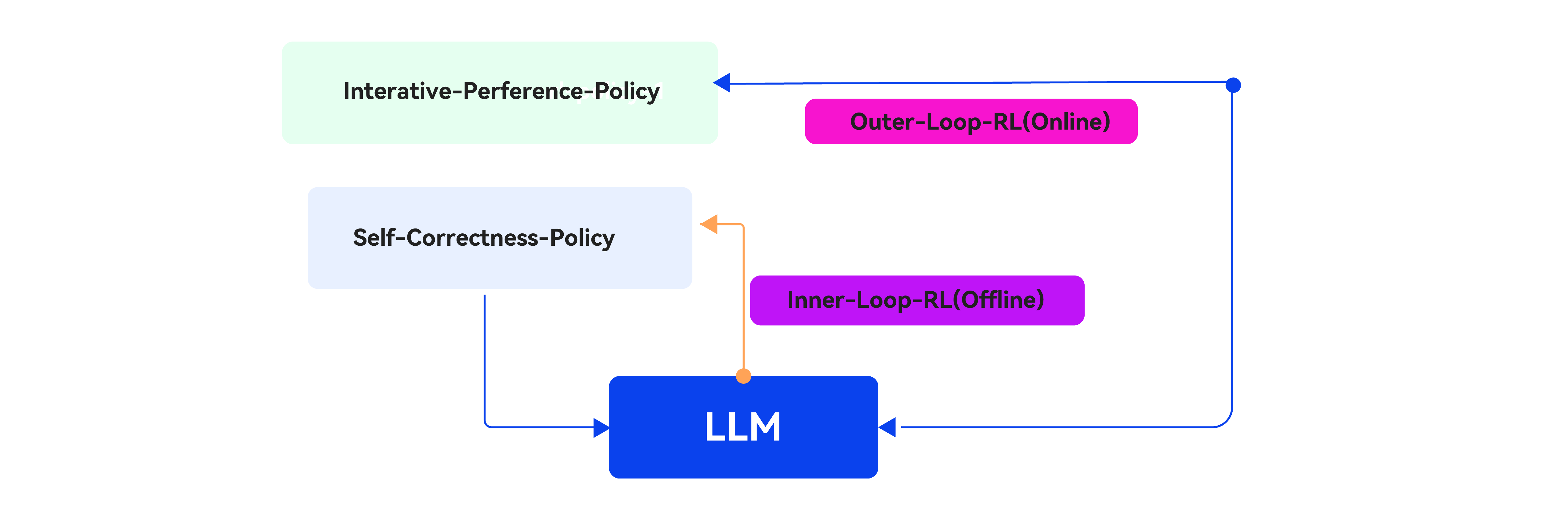}
    \caption{Overview of Towards Intrinsic Self-Correction Enhancement via Iterative Preference Learning. It's consisted of training two policies, self-correctness-policy in the inner-loop reinforcement learning and outer-loop-policy in the outer-loop reinforcement learning. Here, the purple box denotes the learned policy for the first stage. The pink box denotes the learned policy for the second stage.}
    \label{fig:1}
\end{figure}

However, the current iterations of step-level MCTS, LLMs, and RL techniques lack robust self-correction verification. In order to enhance the ability of self-correction of large langugae models(LLM), we therefore propose an improvement to the current state-of-the-art step-level MCTS-DPO  \cite{xie2024monte} method. Specifically, we enhance the self-correction ability of LLMs via reinforcement learning (Figure \ref{fig:1}) to play as the reward model of themselves in the baseline framework \cite{xie2024monte}. Besides, our approach outperforms the OpenMath-Llama-3.1-8B\cite{toshniwal2024openmathinstruct}, dart-math-mistral-7b-uniform \cite{tong2024dart} on MATH \cite{hendrycks2021measuring} with increases in accuracy to 71.34\%(+4.18\%) and 48.06\%(+4.94\%) and LLama-3.1-8B-Instruct \cite{llama3_1},  Mistral-7B-Instruct-v0.1 \cite{yu2023metamath} on GSM8k \cite{cobbe2021training} with increase in accuracy to 86.76\%(+2.00\%) and 38.06\%(+2.28\%)    

\section{Related Works}
Among the current works aimed at enhancing reasoning abilities in LLMs, the following are the most closely related.

\subsection{Self-correcting LLMs}
Previous studies examine self-correction in LLMs across various assumptions and problem settings. Ground-truth answers are applied \cite{kim2024language, shinn2023reflexion} during self-correction. Weak prompts \cite{madaan2024self} are applied for initial responses overestimating the total improvement possible. Then, access to a reward function for evaluating model are used to generate outputs \cite{akyurek2023rl4f, welleck2022generating, zhang2024small, qu2024recursive}. Separate models are trained to perform correction \cite{havrilla2024glore,welleck2022generating, akyurek2023rl4f, paul2023refiner}. Finally, the intersection of LLMs and multi-turn RL builds machinery for optimizing rewards with value-based \cite{farebrother2024stop,shani2024multi,snell2022offline,zhou2024archer}, policy-based \cite{shao2024deepseekmath,xiong2024building}, and model-based \cite{hong2024reference}(self-correct) approaches. However, there has been little exploration of step-level integration between self-correcting LLMs and MCTS, for the enhancement of reasoning in intermediate steps through reinforcement learning.

\begin{table*}
 \centering
 \caption{Based on the Llama-3.1-8B-Instruct and Mistral-7B-Instruct  models, the performance of our method and other methods on GSM8K.}
 \label{tab:1}
\begin{tabular}{cccc}
\hline
Base Model&Approach  &Acc(\%)\\
\hline
&Baseline &84.76& \\
Llama-3.1-8B-Instruct& Intrinsic Self-Correct&86.05&\\
&Base Model + MCTS-DPO&85.67&\\
&Ours&\textbf{86.76}&\\
\hline
&Baseline &35.78& \\
Mistral-7B-Instruct& Intrinsic Self-Correct&37.45&\\
&Base Model + MCTS-DPO&35.71&\\
&Ours&\textbf{38.06}&\\
\hline
\end{tabular}
\end{table*}

\subsection{Fusion of MCTS and LLMs}
Current researches on combining MCTS and LLMs offer benefits for enhancing reasoning capabilities. The integration of MCTS as a policy improvement operator is built to transform the current policy into an improved policy \cite{grill2020monte}. Integrating MCTS into the iterative policy development process could lead to substantial progress in LLMs, especially in areas such as reasoning and decision-making that align with human-like preferences\cite{zhou2024archer, hao2023reasoning}. The instance-level approach relies on sparse supervision, which may overlook crucial information and fail to fully capitalize on MCTS's potential to enhance LLMs \cite{wu2023fine}. And the reliance of MCTS on a critic or a learned reward function may lead to  possible incorrect information \cite{liu2023making}. Many researches have suggested that stepwise evaluations are superior to instance-level evaluations \cite{lightman2023let,li2023making,xie2023decomposition,yao2024tree,rafailov2024direct}. However, the self-evaluation applied in the above integration is not strong. We therefore enhance the ability of self-evaluation in the iteration via reinforcement learning. 

\begin{figure}
    \centering
    \includegraphics[width=1.0\linewidth]{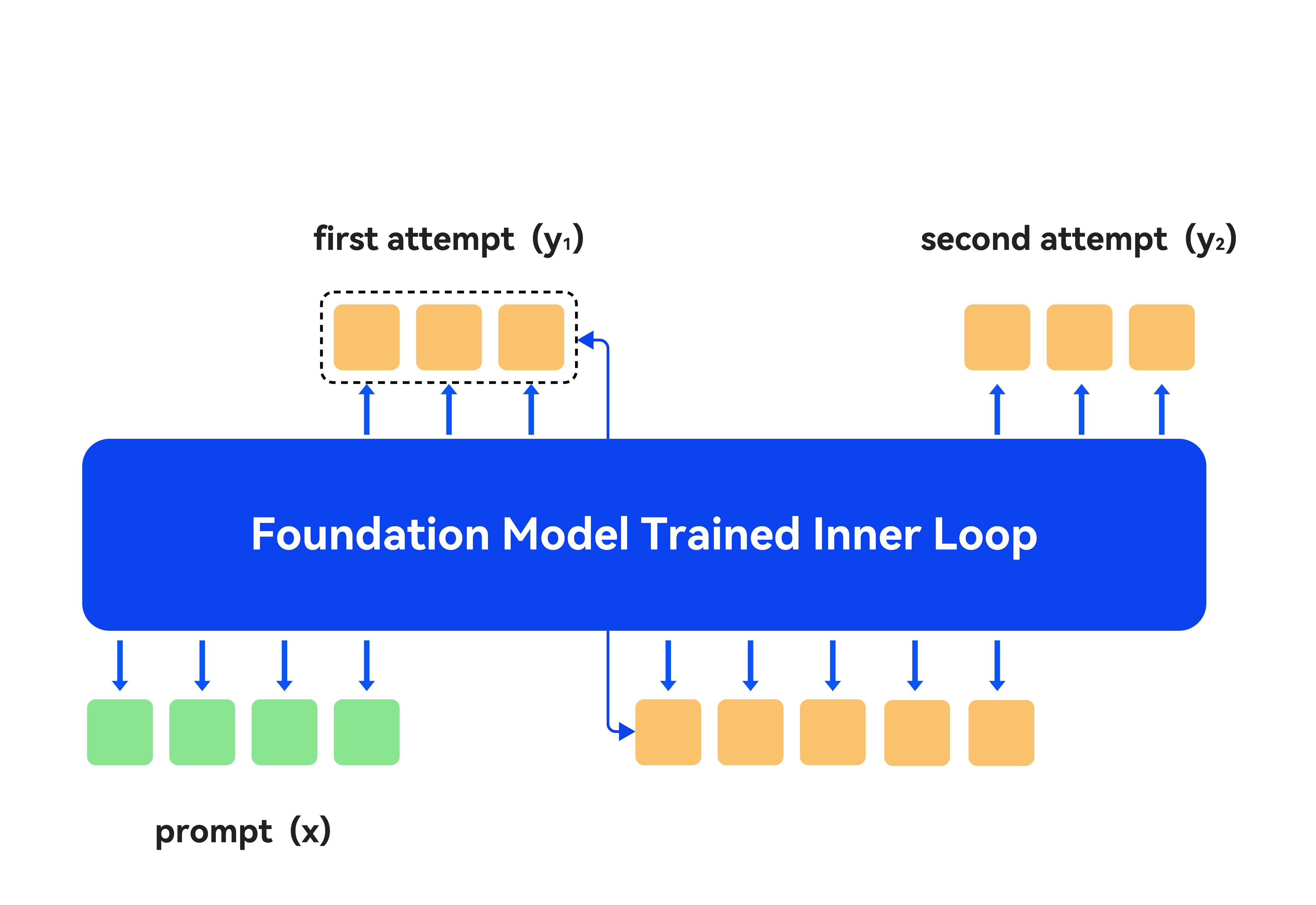}
    \caption{Towards intrinsic Self-Correct LLM in the Inner Loop (Stage I). Here, the green box denotes the input prompt for the LLM at the first stage. The orange box denotes the respondence of the first attempt given the prompt as the input. Then, for the second attempt, the large language model receives the respondence of the first attempt together with the green prompt as input and produces the response of the second attempt(orange box).}
    \label{fig:2}
\end{figure}

\section{Methodology}
We therefore propose a two-stage framework for the self-correction enhancement on the current state-of-the-art step-level MCTS-LLM-DPO \cite{xie2024monte}. At the first stage, we enhance the ability of self-correction of LLMs via self-generated data without any external feedback. At the second stage, we employ the enhanced LLM for the verification enhancement in the step-level preference learning. 

%2 equations
%0.5 Page
\subsection{Stage-I: Intrinsic Data Generation Based Self-Correct LLM} 
In the first stage, our objective is to train LLMs to refine their predictions using solely self-generated data within the towards intrinsic self-correction framework, where models aim to improve their initial responses. \cite{kumar2024training}. 

Concretely, given a dataset $D=\{(x_{i},y_{i}^{*})\}_{i=1}^{N}$ of problems $x_{i}$ and response $y_{i}^{*}$, an LLM policy $\pi_{\theta}(\cdot|[x,\hat{y}_{1:l},p_{1:l}])$ that, given the problem $x$, previous $l$ model attempts $\hat{y}_{1:l}$ at the problem, and auxiliary instructions $p_{1:l}$(eg, instruction to find a mistake and improve the response), solves the problem $x$ as correctly as possible. This formalism is akin to the multi-turn MDP in \cite{qu2024recursive}. We also assume access to an oracle reward $\hat{r}(y,y^{*})$, such as an answer checker \cite{uesato2022solving}, that evaluates the correctness of response $y$ by comparing it with the oracle response $y^{*}$.

At this stage, we aim to find an LLM policy $\pi(\square \mid \circ)$ mapping input tokens $\circ$ to output tokens $\square$ that maximizes the correctness reward obtained from the vertifier at the end of $l+1$ turns. Formally:

\begin{figure}
    \centering
    \includegraphics[width=1.35\linewidth]{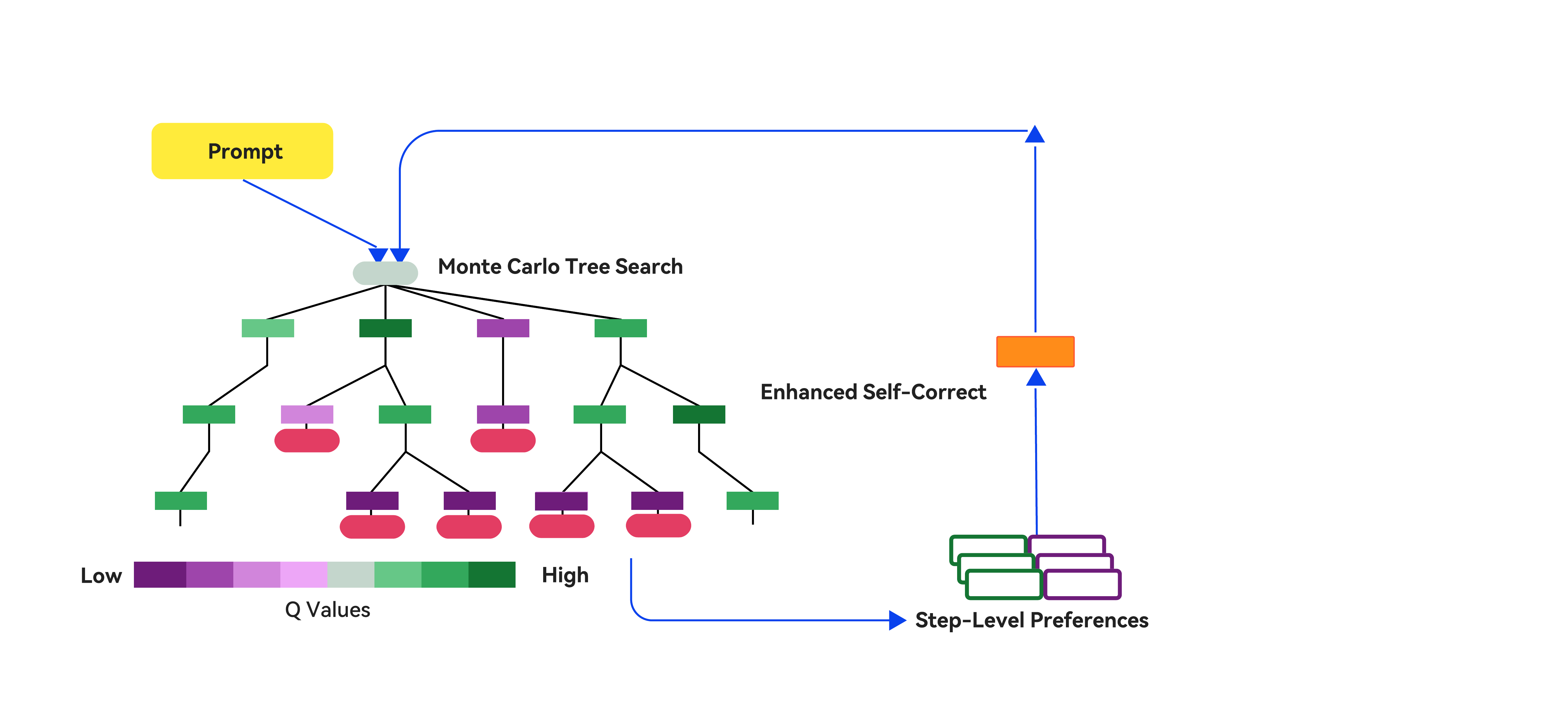}
    \caption{Step-wise Iterative Preference Learning in the Outer Loop. The red circle denotes the termination, the square denotes the intermediate node (Stage II). The orange box denotes the policy learned at the first stage. The Monte Carlo Tree and the step-level preference learning both represent two parts of the iterative preference learning via boosted MCTS.}.
    \label{fig:3}
\end{figure}

\begin{equation}
    \max_{\pi_{\theta_{1}}} \xi_{x,y^{*}\sim \mathcal{D},\hat{y}_{l+1}\sim \pi_{\theta_{1}}(\cdot|[x,\hat{y}_{1:l},p_{1:l}])}[\sum_{i=1}^{l+1}\hat{r}(\hat{y}_{i},y^{*})]
\end{equation}

$\pi_{\theta_{1}}$ is trained over multiple attempts simultaneously, where intermediate turns are supervised indirectly to maximize the sum. We apply a REINFORCE policy gradient training approach \cite{ahmadian2024back} with a KL-divergence penalty against a fix model. 

%2 equations
%0.5 Page
\subsection{Stage-II:Step-level Iterative Preference Learning}
At the second stage, we employ step-level iterative preference learning with enhanced self-correction LLM (achieved at the first stage) to enhance the step-wise verification via self-correction. 

We define the state at step $t$, $s_{t}$ as the prefix of the reasoning chain, $a$ as the corresponding action, with the addition of a new reasoning step transitioning the state to $s_{t+1}$, where $s_{t+1}$ is the concatenation of $s_{t}$ and $a$. Utilizing the model’s current policy $\pi_{\theta_{2}}$, we sample candidate steps from its probability distribution $\pi_{\theta_{2}}(a \mid x, s_{t})$, with $x$ being the task's input prompt. 

Then, the MCTS process begins from a root node, $s_{0}$, as the sentence start or incomplete response, and unfolds in four iterative stages: selection, expansion, enhanced-self-verify, and backup. The four details are represented as the following:

\paragraph{Select}
The objective of this phase \cite{xie2024monte} is to identify nodes that balance search quality and computational efficiency. The selection is guided by two key variables: $Q(s_{t},a)$, the value of taking action $a$ in state $s_{t}$ , and $N(s_{t})$, the visitation frequency of state $s_{t}$. These variables are crucial for updating the search strategy, as explained in the backup section. To navigate the trade-off between exploring new nodes and exploiting visited ones, we employ the Predictor + Upper Confidence bounds applied to Trees (PUCT) \cite{rosin2011multi}. At node $s_{t}$, the choice of the subsequent node follows the formula:

\begin{equation}
    s_{t+1}^{*} = \arg \max_{s_{t}}[\mathcal{Q}(s_{t},a) + c_{puct} \cdot p(a \mid s_{t}) ]\frac{\sqrt{N(s_{t})}}{1+N(s_{t+1})}
\end{equation}
where $p(a \mid s_{t}) = \pi_{\theta_{2}}(a \mid x, s_{t})/|a|^{\lambda}$ denotes the policy $\pi_{\theta_{2}}$'s probability distribution for generating a step $a$, adjusted by a $\lambda$-weighted length penalty to prevent overly long reasoning chains.

\begin{table*}[htbp]
 \centering
 \caption{Based on the OpenMath2-Llama3.1-8B  and dart-math-mistral-7b-uniform models, the performance of our method and other methods on MATH}
 \label{tab:2}

\begin{tabular}{cccc}
\hline
Base Model&Approach  &Acc(\%)\\
\hline
&Baseline &67.16& \\
OpenMath2-Llama3.1-8B&Intrinsic Self-Correct&69.78&\\
&Base Model + MCTS-DPO&68.06&\\
&Ours&\textbf{71.34}\\
\hline
&Baseline &43.42& \\

dart-math-mistral-7b-uniform&Intrinsic Self-Correct&44.50&\\
&Base Model + MCTS-DPO&45.56&\\
&Ours&\textbf{48.06}&\\
\hline
\end{tabular}
\end{table*}

\paragraph{Expand}
Expansion occurs at a leaf node during the selection process to integrate new nodes and access rewards \cite{xie2024monte}. The reward $r(s_{t},a)$ for executing step $a$ in state $s_{t}$ is quantified by the reward difference between $R(s_{t})$ and $R(s_{t+1})$, highlighting the advantage of action $a$ at $s_{t}$. As defined below, reward computation merges outcome correctness $\mathcal{O}$ with self-evaluation $\mathcal{C}$. We assign the outcome correctness to be $1$, $-1$ and $0$ for correct terminal, incorrect terminal, and unfinished intermediate states, respectively. Following \cite{xie2023decomposition}, we define self-evaluation as the following equation, where $A$ denotes the confidence score in token-level probability for the option indicating correctness. $prompt_{eval}$ denotes the prompt applied by the verified LLM. Future rewards are anticipated by simulating upcoming scenarios through roll-outs, following the selection and expansion process until reaching a terminal state. 

\begin{equation}
    \mathcal{R}(s_{t}) = \mathcal{O}(s_{t}) + \mathcal{C}(s_{t})
\end{equation}

\begin{equation}
    \mathcal{C}(s_{t}) = \pi_{\theta_{2}}(A \mid prompt_{eval},x,s_{t})
\end{equation}

\paragraph{Enhanced-Self-Verify}
After each expanding step, we conduct a self-correct to further correct possible misleading or even wrong reasoning generated intermediate outcome, the self-correction is provided from the same LLM achieving the policy $\pi_{\theta_{1}}$ at the first stage. 

\begin{table*}[htbp]
 \centering
 \caption{Ablation experiments: the performance of different models to initialize the policy model and reward model on MATH. }
 \label{tab:pagenum}
\begin{tabular}{cccc}
\hline
Policy model& Reference model& Reward model&Acc(\%) \\
\hline
OpenMath2-Llama3.1-8B& OpenMath2-Llama3.1-8B& OpenMath2-Llama3.1-8B& 68.06\\
OpenMath2-Llama3.1-8B& OpenMath2-Llama3.1-8B& SPL-Model&67.06 \\
SPL-Model & SPL-Model& SPL-Model& \textbf{71.34}\\
ISC-Model & ISC-Model& SPL-Model& 65.34\\
\hline
dart-math-mistral-7b-uniform& dart-math-mistral-7b-uniform& dart-math-mistral-7b-uniform& 45.56\\
dart-math-mistral-7b-uniform& dart-math-mistral-7b-uniform& SPL-Model& 46.84\\
SPL-Model & SPL-Model& SPL-Model& \textbf{48.06}\\
ISC-Model & ISC-Model&SPL-Model& 47.22\\
\hline
\end{tabular}
\end{table*}

\begin{equation}
    \mathcal{R}(s_{t}^{correct}) = \mathcal{R}(s_{t}) + \hat{\mathcal{C}}(s_{t})
\end{equation}

\begin{equation}
    %\mathcal{R}(s_{t}^{correct}) = \pi_{\theta_{1}}(\hat{\mathcal{C}}(s_{t}) \mid (\mathcal{R}(s_{t}),prompt_{eval},x,s_{t}))\\
    \hat{\mathcal{C}}(s_{t})^{stage_{1}} = \pi_{\theta_{1}}(A|prompt_{eval},x,s_{t})
\end{equation}

\paragraph{Backup}
Once a terminal state is reached, we carry out a bottom-up update from the terminal node back to the root. We update the visit count $N$, the state value $V$ and the transition value $Q$ which remains the same.

\section{Experiments}
We evaluate the effectiveness of DPO by MCTS and iterative preference learning on GSM8K and MATH reasoning tasks. We use the Llama-3.1-8B-Instruct\cite{dubey2024llama}, Mistral-7B-Instruct-v0.1 \cite{jiang2023mistral} as the base model on GSM8K \cite{cobbe2021training} and OpenMath2-Llama3.1-8B \cite{toshniwal2024openmath2}, dart-math-mistral-7b-uniform\cite{tong2024dart} as the base  model on MATH \cite{hendrycks2021measuring}. Iterative Preference Learning\cite{kumar2024training} combined with online MCTS-DPO\cite{xie2024monte} work well.

\subsection{Datasets}
We aim to demonstrate the effectiveness and versatility of our approach by focusing on arithmetic reasoning. We utilized two datasets: GSM8K \cite{cobbe2021training}, which consists of grade school math word problems, and MATH \cite{hendrycks2021measuring}, featuring challenging competition math problems. 

\subsection{Main Results}
Our results on GSM8K and MATH are shown in Table \ref{tab:1} and Table \ref{tab:2}. Our method achieved increases on accuracy across all four LLMs and two datasets. With towards intrinsic self-correct training, we obtained a model capable of self-correction. We further boosted LLMs performance on these datasets through iterative preference learning. Specifically, on GSM8K, we achieved 2.00\% and 2.28\% increases in  accuracy for Llama-3.1-8B-Instruct and Mistral-7B-Instruct respectively. On MATH, we observed improvements of 4.18\% and 4.64\% for OpenMath2-Llama3.1-8B and dart-math-mistral-7b. All of these results demonstrate the effectiveness of our approach in enhancing LLM performance across datasets with various difficulties.

%Our results on GSM8K and MATH are shown in Table 1 and Table 2, correspondingly. Our method achieved increases on accuracy across all four LLMs and two datasets. With intrinsic self-correct training, we achieved increases in accuracy of 1.29\% and 3.80\% for Llama-3.1-8B-Instruct and Mistral-7B-Instruct on GSM8K, 2.62\% and 1.08\% for OpenMath2-Llama3.1-8B and dart math-mistral-7b on MATH. With the trained model capable of self-correction, we further boosted LLMs performance on these datasets through iterative preference learning. On GSM8K, we achieved 0.71\% and 0.61\% accuracy relative to stage I for Llama-3.1-8B-Instruct and Mistral-7B-Instruct. On MATH, we observed improvements of 1.56\% and 3.56\% for OpenMath2-Llama3.1-8B and dart-math-mistral-7b. All of these results demonstrate the effectiveness of our approach in enhancing LLM performance across datasets with various difficulties.
\subsection{Ablation Studies}
To demonstrate the effectiveness of Towards Intrinsic Self-Correct and Step-level Iterative Preference Learning, we need to verify that neither method alone can achieve the same performance as their combination. Since GSM8K is relatively simple and easier for models to learn, its persuasiveness as a test case is limited. Therefore, we conducted our verification on the more challenging MATH dataset. For clarity, we denote the models generated by Intrinsic Self-Correct and Step-level Iterative Preference Learning as ISC-Model and SPL-Model, respectively. As shown in Table\ref{tab:pagenum}, the results indicate that the combination where both the policy model and the reward model are SPL-Model consistently outperforms other configurations. For OpenMath2-Llama3.1-8B, all SPL-Model setting achieves 71.34\% in accuracy surpasses other configuration by 3.28\% to 6.00\%. Similarly, for dart-math-mistral-7b-uniform, all SPL-Model configurations consistently outperform other settings by 0.84\% to 2.50\%. All of these results  demonstrate the superior synergy of the two methods.

% In order to prove the effectiveness of Intrinsic Self-Correct and Step-level Iterative Preference Learning, we need to verify that a single Intrinsic Self-Correct or Step-level Iterative Preference Learning method cannot achieve the effect of combining the two. Since the GSM8K data is relatively simple and the model is easy to learn, the persuasiveness is not strong, so we verify it on the difficult MATH dataset. For the sake of convenience, the models generated by Intrinsic Self-Correct and step-level Iterative Preference Learning are named ISC-Model and SPL-Model respectively. As shown in the table\ref{tab:pagenum}, we find that when both the policy model and the reward model are SPL-Model, they are better than other combinations.

\section{Discussion}
We start our step-level and self-correct level learning for enhancing the ability of reasoning for LLMs. At the first stage, we integrate an enhanced self-correct model via self-supervised learning with step-level preference learning. Then, we are going to begin our second stage which improves the step-level preference and self-correction via online reinforcement learning. Finally, we are planning to conduct experiments on other reasoning dataset. 

\bigskip

\bibliography{aaai25}

\end{document}